\documentclass{article}

\usepackage{spconf,amsmath,epsfig}
\usepackage{url}
\usepackage{ifthen}

\begin{document}\sloppy
\def\x{{\mathbf x}}
\def\L{{\cal L}}

\title{Perspective-aware Convolution for Monocular 3D object detection}
\name{$^1$Jia-Quan Yu, $^2$Soo-Chang Pei}
\address{$^1$Graduate Institute of Communication Engineering,\\
National Taiwan University, Taiwan\\
E-mail: kenyu910645@gmail.com\\
$^2$Department of Electrical Engineering,\\
National Taiwan University, Taiwan\\
E-mail: peisc@ntu.edu.tw}
\maketitle

\begin{abstract}
Monocular 3D object detection is a crucial and challenging task for autonomous driving vehicle, while it uses only a single camera image to infer 3D objects in the scene. To address the difficulty of predicting depth using only pictorial clue, we propose a novel perspective-aware convolutional layer that captures long-range dependencies in images. By enforcing convolutional kernels to extract features along the depth axis of every image pixel, we incorporates perspective information into network architecture. We integrate our perspective-aware convolutional layer into a 3D object detector and demonstrate improved performance on the KITTI3D dataset, achieving a 23.9\% average precision in the easy benchmark. These results underscore the importance of modeling scene clues for accurate depth inference and highlight the benefits of incorporating scene structure in network design. Our perspective-aware convolutional layer has the potential to enhance object detection accuracy by providing more precise and context-aware feature extraction.
\end{abstract}

\begin{keywords}
Dilation Convolution, Monocular 3D Object Detection, Perspective-aware
\end{keywords}

\section{Introduction}
\label{sec:intro}

Estimating object depth and understanding the scene structure are critical tasks in image recognition, especially in the context of autonomous driving where safety is paramount. While inferring object depth from a single camera image is challenging, existing approaches predominantly rely on costly active sensors like LiDAR, Radar, or infrared cameras that directly provide depth information. However, cameras offer a more cost-effective and practical alternative, given their ease of installation on vehicles. The main hurdle with cameras is the absence of depth information in 2D images, posing a significant challenge for depth estimation algorithms.

\begin{figure}[t]
\begin{center}
\includegraphics[width=8.5cm]{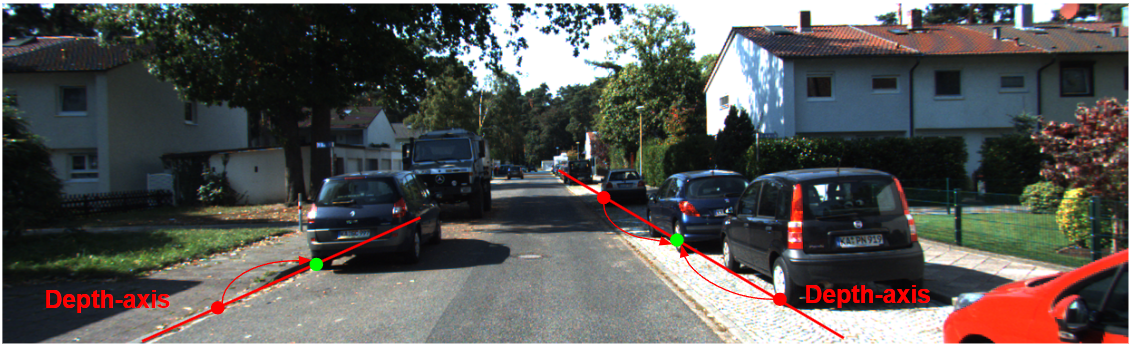}
\end{center}
\caption{Illustration of long-range dependency that aids in depth inference. Directly predicting the depth of the green dot is challenging. However, if we can determine the depth-axis for every pixel and extract the surrounding front and back pixels along this axis, it can significantly enhance the accuracy of predicting the depth for the green dot. To achieve this, we utilize the camera intrinsic matrix to derive the depth-axis and introduce a skewed convolutional kernel designed to capture features along this axis.}
\label{fig:long-range}
\end{figure}

Despite the absence of depth information in camera images, we posit that the human perception system is capable of inferring depth from limited visual cues by leveraging other scene information. For instance, as depicted in Fig.\ref{fig:long-range}, humans can infer the location of objects based on the relative positions of other nearby objects in the scene, exploiting their understanding of scene structure. By comparing the distances between adjacent objects, depth can be estimated even in cases where occlusion occurs. Hence, we believe understanding scene structure plays a pivotal role in accurate depth estimation. Our objective is to integrate this perspective information into our convolutional neural network.

In this paper, we present a novel approach called perspective-aware convolution (PAC) to enhance the capability of convolutional neural networks in capturing perspective-related features. PAC extracts feature along the depth axis by adjusting the shape of the convolutional kernels. Additionally, we introduce a PAC module that integrates multiple dilation rates within parallel convolutional branches. By incorporating the PAC module into 3D object detection networks, we enable them to generate perspective-aware feature maps, enhancing their ability to analyze objects in specific perspective scene structure.

Finally, to demonstrate the effectiveness of the PAC module, we evaluate it in KITTI 3D object detection challenge, where objects are defined as cuboids in the camera coordinate system. We train and evaluate our network on the KITTI dataset and achieve 23.53\% AP on easy metric.

The remainder of this paper is organized as follows: Section 2 introduces the related work on 3D object detection and dilated convolution modules. Section 3 explains our proposed perspective-aware convolution. Subsequently, we report our experimental results on the KITTI dataset in Section 4.
\section{Related Works}

\subsection{Convolutional Module}

Convolutional layers are essential components of deep learning network. It plays a crucial role in extracting features from image by convolving a learnable kernel with the image pixel in a sliding-window fashion. While convolutional layers excel at recognizing local patterns, they have limitations when it comes to capturing long-range dependencies within an image. To address this limitation, researchers have explored various techniques to increase network's receptive field, which is, the size of the image region that the network considers when making predictions at a particular location.

To enlarge the receptive field of a network, researchers often resort to building deeper networks or increasing number of downsampling to expand the receptive field size; however, these approaches can result in the loss of fine-grained details in feature maps. Alternatively, dilated convolutions, as introduced in \cite{dilationConv}, provide another solution. By skipping a certain number of image pixels during convolution, determined by the dilation rate, dilated convolutions enable the network to have larger receptive fields while no need of down-sampling. Expanding on the benefits of dilated convolutions, DeepLabv2\cite{aspp} introduced atrous spatial pyramid pooling (ASPP) to further enhance feature extraction. The ASPP module incorporates multiple parallel dilated convolutions with different dilation rates, allowing for the extraction of multi-scale features from the same feature map. Another approach to enhancing the feature extraction capability of convolutional layers is the receptive field block(RFB)\cite{rfb}, which adjusts the kernel size based on the corresponding dilation rate of the convolutional module.

However, all the aforementioned methods rely on fixed kernel shapes that are predefined before the training process. To address this limitation, Dai et al. proposed a novel convolutional layer called deformable convolutional networks (DCN)\cite{dcn}, and its improved version DCNv2\cite{dcnv2}. These convolutional modules enable the network to dynamically adjust the shape of convolutional kernel during training, making the feature extractor more adaptive to the scene structure. While this method introduces more flexibility to the training process, it also incurs a non-negligible overhead.

Despite the extensive research on convolution layers, there are few methods that incorporate perspective information into the network. Therefore, drawing inspiration from the ASPP module and deformable convolution, we propose our perspective-aware convolution (PAC) module, which adjusts the kernel shape based on the depth axis in the image. This novel module allows the network to capture the underlying scene geometry and perspective, enhancing its ability to understand the 3D structure of the environment. In Section 3, we will provide a detailed explanation of our PAC module and its integration within our proposed method.

\subsection{Monocular 3D Object Detection}
Monocular 3D object detection is a rapidly evolving research field that utilize single camera images as input to estimate the 3D appearance of objects within the scene. In this section, we provide an overview of the fundamental concepts of 3D object detection and discuss some relevant prior works in the field.

\begin{figure}[t]
\begin{center}
\includegraphics[width=8.5cm]{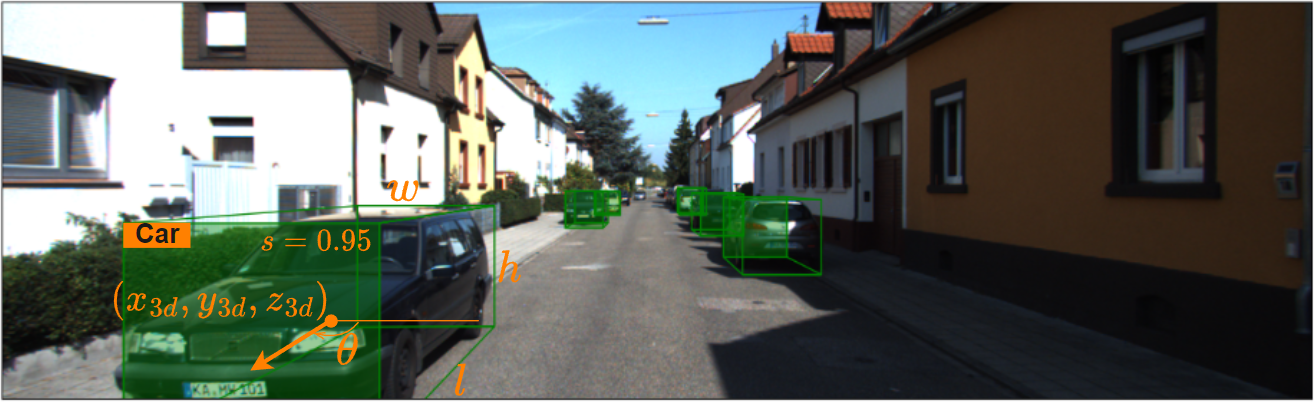}
\end{center}
\caption{3D object detection. The cuboids in defined in camera coordinate and we need to find the location $(x, y, z)$, dimension $(w, h, l)$, and orientation $\theta$ of each cuboid. }
\label{fig:3d_od}
\end{figure}

In the field of 3D object detection, an object is represented by a cuboid which respect to camera coordinates. This cuboid is characterized by seven parameters: centroid coordinate $(x, y, z)$, which specifies the location of the cuboid center relative to the camera center, and dimensions $(w, h, l)$, which correspond to the width, height, and length of the cuboid, respectively. Additionally, the yaw angle $\theta$ denotes the orientation of the cuboid. Our objective in 3D object detection is to identify objects within images and accurately localize them in camera coordinates by predicting these seven variables $(x, y, z, w, h, l, \theta)$ for each object. This concept is illustrated in Fig\ref{fig:3d_od}.

\begin{figure}[t]
\begin{center}
\includegraphics[width=8.5cm]{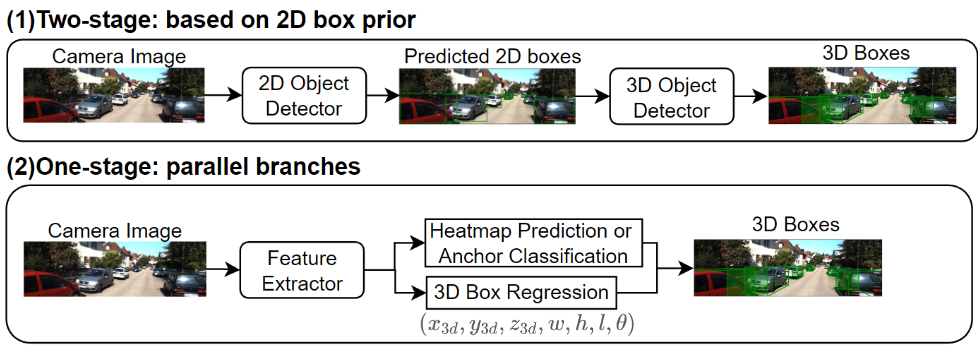}
\end{center}
\caption{
Related work in 3D object detection can be categorized into two-stage and one-stage detectors. Two-stage detectors utilize a predicted 2D bounding box as a prior and extract features within the 2D box, whereas one-stage detectors treat the prediction of 2D and 3D boxes as a unified task, predicting both boxes in parallel branches.}
\label{fig:3dod_one_two_stage}
\end{figure}

The related work in 3D object detection can be divided into two main branches, as we shown in Fig\ref{fig:3dod_one_two_stage}: two-stage detectors and one-stage detectors. Two-stage detectors, such as Deep3DBox\cite{deep3dbox} and FQNet\cite{fqnet}, rely on a predicted 2D bounding box as prior information. These methods assume that all projected corners of a 3D bounding box must lie within its corresponding 2D bounding box. While these approaches reduce the complexity of the problem, they are sensitive to inaccuracies in the predicted 2D bounding box. On the other hand, other two-stage detector, including MonoDIS\cite{monodis}, ROI10D\cite{roi10d}, MonoGRNet\cite{monogrnet}, and MonoPSR\cite{monopsr}, utilize the predicted 2D bounding box as a region proposal to extract features and predict 3D box geometry from it. In specific, MonoDIS uses RoIAlign to extract fixed-length features from the regions of interest (ROIs) and proposes a disentangle loss function to avoid interference between each loss term and aid in faster convergence. MonoGRNet employs both early and deep features in the backbone network and uses deep features only for tasks that require a higher receptive field, such as depth estimation, to prevent excessive downsampling of the feature map. ROI10D and MonoPSR both utilize a pre-trained depth estimation model to improve their accuracy in estimating object depth.

One-stage detectors take a different approach compared to two-stage detectors as they do not rely on 2D box priors. Instead, they treat 3D objects as extensions of 2D objects and utilize a unified network to predict both 2D and 3D bounding boxes in parallel branches without the need for region-specific feature extraction. One example is M3D-RPN\cite{m3drpn}, which reformulates the 2D detector network to capture 3D proposals using a shared network for both tasks. M3D-RPN also incorporates statistical data from the training set to determine the 3D anchor box prior. Additionally, M3D-RPN proposes a depth-aware convolution, which employs separate kernels to extract features from different image rows, enabling the network to handle depth features separately. Another method, Ground-aware\cite{gac}, focuses on locating and extracting the ground-contact point of each object to enhance its depth perception capability.

Another example of a one-stage detector is the keypoint-based network, as proposed by CenterNet\cite{centernet}. By detecting keypoints such as the center point or corner points of the 3D objects, these methods can efficiently predict 3D bounding boxes. RTM3D\cite{rtm3d} takes a similar approach by identifying eight corners and the object center of the cuboid. It also introduces a feature pyramid network to capture multi-scale keypoints. SMOKE\cite{smoke} simplifies the network by eliminating the 2D box regression branch and incorporates MonoDIS's disentangle loss to improve convergence. MonoPair\cite{monopair} focuses on leveraging the relationship between adjacent objects by predicting keypoints at the midpoint of each adjacent object pair. Additionally, MonoPair introduces uncertainty in the regression branch and utilizes it for post-processing optimization of the detection results. MonoFlex\cite{monoflex} addresses truncated objects whose centers lie outside the image by proposing an edge fusion module to separate feature learning from truncated object prediction. MonoFlex employs an ensemble approach for depth prediction, considering depth uncertainty and the predicted 3D bounding box's multiple pixel heights to enhance depth estimation accuracy.

Overall, two-stage detectors have been pioneering in the field of 3D object detection, using 2D bounding boxes as priors. However, they tend to have performance degradation when the 2D bounding box predictions are inaccurate. On the other hand, one-stage detectors offer a more unified network architecture with improved performance. Therefore, in our work, we choose to adopt an anchor-based one-stage detector for our experiments, leveraging its superior performance in 3D object detection tasks.

\section{Proposed Method}

\begin{figure}[ht!]
  \centering
  \includegraphics[width=8.5cm]{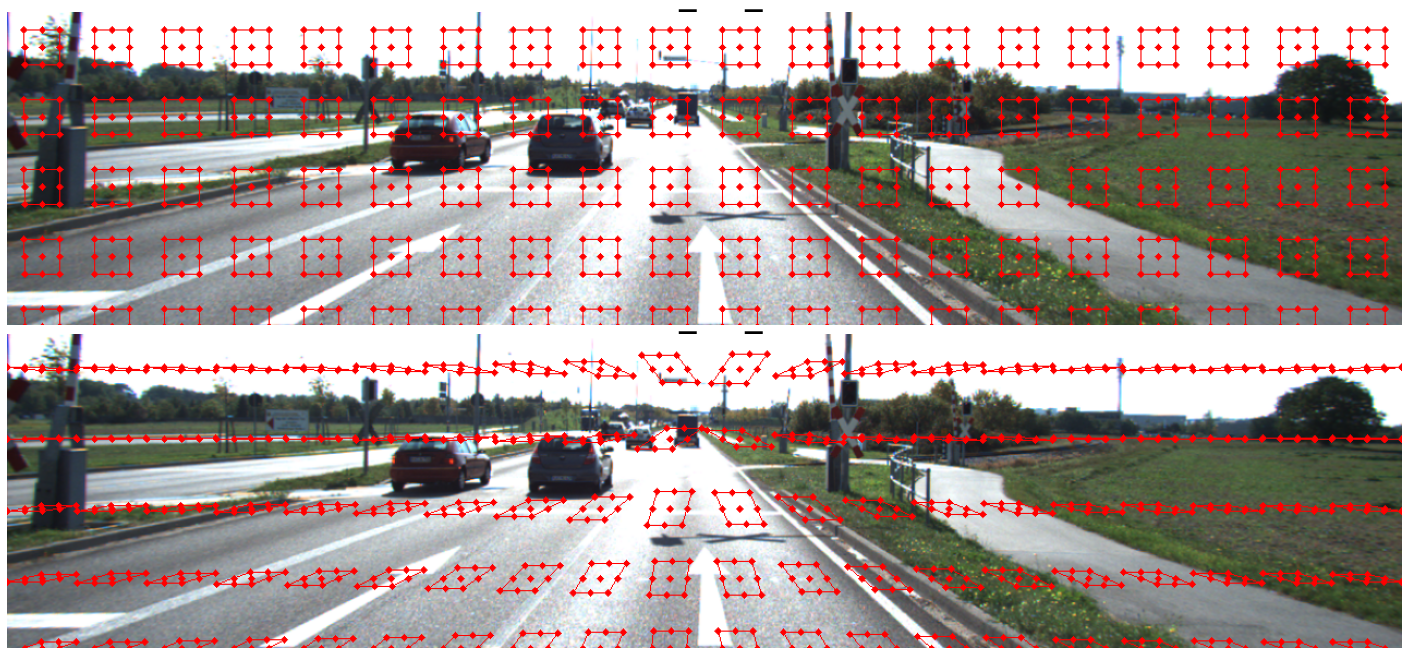}
  \caption{Perspective-aware convolution. The red square in the image represent the kernel shape. The top image illustrates the kernel shape of dilation convolution, while the bottom image demonstrates the kernel shape of perspective-aware convolution. Our perspective-aware kernel dynamically adjusts its shape based on the depth axis at each pixel.}
  \label{fig:pac_example_2}
\end{figure}

\section{Perspective-aware Convolution}
We propose a novel perspective-aware convolutional layer that extracts features along the perspective lines at each pixel location. Our motivation arises from the recognition that the depth-axis-adjacent objects contain essential information for object depth estimation. Conventional convolutional layers often struggle to capture long-range dependencies in images. To overcome this limitation, we attempt to explicitly inject the perspective information into the network by guiding the convolutional kernels to capture features along the depth axis of each pixel. These axis are straight lines parallel to the camera's depth axis in the camera coordinate system. By projecting these lines onto the image plane using the camera pinhole model, we obtain an indication of how each pixel would move on the image if its depth value were to change. Additionally, we can derive the angle between the depth axis and the u-axis, which we refer to as the perspective angle, and use it to represent the perspective line for simplicity. In the following part of this section, we will elaborate on how we derive the perspective angle based on the camera pinhole model.

To get the perspective angle, we begin with the formulation of the pinhole camera model:
\begin{equation}
\label{eq:pinhole_model}
\begin{bmatrix}u\\v\\1\end{bmatrix}=
    \frac{1}{Z_w}
   \begin{bmatrix}
     f_x & 0   & C_x\\
     0   & f_y & C_y\\
     0   & 0   & 1  \\
   \end{bmatrix}
   \begin{bmatrix}X_w \\Y_w\\Z_w \end{bmatrix}
\end{equation}

In Equation \ref{eq:pinhole_model}, $(X_w, Y_w, Z_w)$ represents the coordinates of a point in the camera coordinate system, while $(u, v)$ represents the projected pixel coordinates on the image plane. The parameters $f_x$ and $f_y$ denote the focal lengths expressed in pixel units for the u and v axes. The values $C_x$ and $C_y$ correspond to the principal point, which is the intersection point of the optical axis with the image plane.

To determine the amount of pixel displacement caused by a change in depth, we differentiate Equation \ref{eq:pinhole_model} with respect to $Z_w$ and obtain:

\begin{equation}
\label{eq:derivative}
    \begin{cases}
    \vspace{0.5em}
    \displaystyle \frac{du}{dZ_w} = -\frac{X_wf_x}{Z_w^2} \\
    \displaystyle \frac{dv}{dZ_w} = -\frac{Y_wf_y}{Z_w^2} \\
    \end{cases}
\end{equation}

where $(X_w,Y_w,Z_w)$ represents a 3D point in camera coordinates, which is inversely projected from a pixel $(u, v)$. Since the depth of each pixel is unknown, we assume that all image pixels lie on the ground plane and set $Y_w$ equal to the height of the ground plane, denoted as $Y_0$. By applying Equation \ref{eq:pinhole_model}, we can inversely project $(u,v)$ back to the camera coordinate system and the inversely projection equation is as following:

\begin{equation}
\label{eq:specific}
    \begin{cases}
    \vspace{0.7em}
    \displaystyle  X_0 = \frac{(u_0-C_x)Y_0f_y}{(v_0-C_y)f_x} \\
    \displaystyle  Y_0 = Y_0\\
    \displaystyle  Z_0 = \frac{Y_0f_y}{v_0-C_y} \\
    \end{cases}
\end{equation}

Here, $(u_0, v_0)$ represents a specific pixel on the image, and $(X_0, Y_0, Z_0)$ represents the inversely projected 3D point in camera coordinates. By substituting $(X_0, Y_0, Z_0)$ into $(X_w, Y_w, Z_w)$ in Equation \ref{eq:derivative}, we can calculate the derivatives and the perspective angle $\phi$ is determined by the following equation:
\begin{equation}
\label{eq:perspective_angle}
    \phi = \operatorname{atan2}\left(\frac{dv}{dZ_w}, \frac{du}{dZ_w}\right)
\end{equation}

Using Equations \ref{eq:derivative} and \ref{eq:perspective_angle} , we can calculate the perspective angle for each image pixel $(u, v)$. This perspective angle allows us to guide kernel to change their shape according to its pixel coordinate and perspective angle, as illustrated in Fig. \ref{fig:pac_example_2}.

\begin{figure*}
  \centering
  \includegraphics[width=18cm]{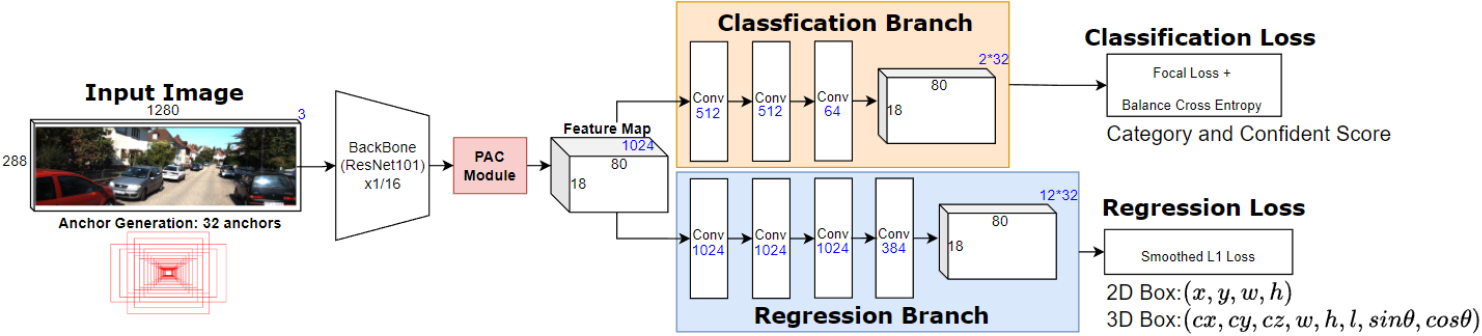}
  \caption{Our proposed architecture for 3D object detection. We adopt the network proposed by Ground-aware\cite{gac}, which is a one-stage anchor-based architecture. We add our PAC module after the feature extractor, which is ResNet-101 in our implementation,  to obtain the perspective-aware feature map. The top branch focuses on classifying positive and negative anchor predictions, while the lower branch is responsible for regressing the geometry of both 2D and 3D bounding boxes. This architecture enables accurate object detection by leveraging the benefits of the PAC module in capturing scene structure and improving 3D box regression.}
  \label{fig:architecture}
\end{figure*}

\subsection{Perspective-aware Convolutional Module}

\begin{figure}[ht!]
  \centering
  \includegraphics[width=8.5cm]{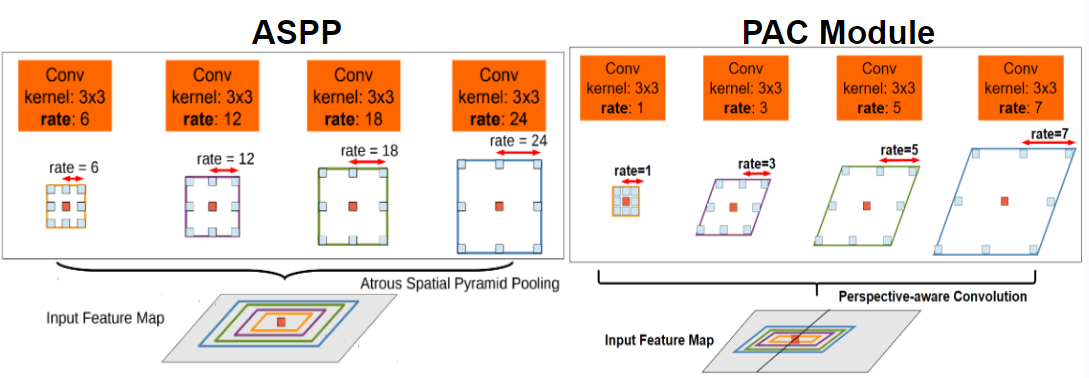}
  \caption{Comparison of PAC module and ASPP module. Both modules utilize parallel branches to capture multi-scale features. However, the key distinction lies in the kernel shape employed for feature extraction. While the ASPP module utilizes a regular kernel shape, the PAC module incorporates a tilted kernel shape to guide feature extraction along the perspective line.}
  \label{fig:pac_and_ASPP}
\end{figure}

In addition to incorporating a single PAC convolution layer, we employ a multiple-branch design inspired by ASPP\cite{aspp}, utilizing different dilation rates for each branch. Our objective is to capture multi-scale features along each perspective line with the PAC module. Furthermore, to ensure the preservation of regular features, we include a branch with a standard 3x3 kernel in our PAC module. This design choice guarantees that the regular feature map passes through our module without any alterations. A comparison between ASPP module and PAC module is depicted in Fig. \ref{fig:pac_and_ASPP}.

\section{Experiments}

In this section, we present the experimental results of our proposed PAC module in the task of 3D object detection. We integrate the PAC module into the baseline network of Ground-aware\cite{gac} and aim to enhance its performance. The training and evaluation of the detector are conducted on the KITTI3D dataset, consisting of 3711 training images and a validation set of 3768 images. We adopt to the data split recommended by Chen et al.\cite{chen_split}. During training, we set the batch size to 8 and utilize the Adam optimizer with a learning rate of $10^{-4}$. To speed up the network, we crop the top 100 pixels from the images and resize them to 288x1280. Additionally, we apply horizontal flipping and photometric distortion techniques to improve the diversity of the training data.

We present our experiment results in Table \ref{tab:compare_with_related_works} where we compare with other 3D object detection networks. Our proposed method surpasses all other detectors in terms of average precision, showing the effectiveness of our PAC module.

\begin{table*}[t]
\begin{center}
\caption{Experimental results of the 3D object detection algorithm on the KITTI3D validation dataset. The best performance in each column is indicated in bold font.} \label{tab:compare_with_related_works}
\begin{tabular}{|c|ccc|ccc|}
  \hline
  &\multicolumn{3}{c|}{Car AP 3D (IoU=0.7)} &\multicolumn{3}{c|}{Car AP BEV (IoU=0.7)} \\
  Methods & Easy & Moderate & Hard & Easy & Moderate & Hard \\
  \hline
  MonoGRNet\cite{monogrnet} & 12.28 &  7.76 &  5.91 & 19.89 & 12.94 & 10.31 \\
  SMOKE\cite{smoke} &  6.96 &  4.30 &  3.98 & 12.73 &  7.93 &  6.94 \\
  DD3D\cite{dd3d} & 19.16 & 15.27 & 13.37 & 25.72 & 20.78 & 18.38 \\
  MonoFlex\cite{monoflex} & 22.14 & 16.19 & 14.18 & 29.30 & 21.91 & 18.82 \\
  Ground-aware\cite{gac} & 21.90 & 16.06 & 13.17 & 28.29 & 20.98 & 17.59 \\
  \textbf{Ours(Ground-aware+PAC Module)} & \textbf{23.53} & \textbf{17.23} & \textbf{14.33} & \textbf{30.57} & \textbf{22.44} & \textbf{19.07} \\
  \hline
\end{tabular}
\end{center}
\end{table*}

Additionally, we compare PAC module with other convolutional module mentioned in Section 2 and report the result in \ref{tab:compare_with_modules}. We adopt Ground-aware network as the baseline and applied the dilation convolution in the last three convolutional layers of the backbone and set the dilation rate to (2,2,2) and (3,3,3) in our experiment. We also add DCN, RFB, ASPP, and PAC after the feature extractor, following their respective recommended settings in each paper. We conducted experiments with a single layer of PAC, where the dilation rate is set to two. As for the PAC module, we set the dilation rate to 2, 4, 6, and 8 in each parallel branch.

\begin{table*}[t]
\begin{center}
\caption{Experimental results of the 3D object detection algorithm on the KITTI3D validation dataset. The best performance in each column is indicated in bold font.} \label{tab:compare_with_modules}
\begin{tabular}{|c|ccc|ccc|}
  \hline
  &\multicolumn{3}{c|}{Car AP 3D (IoU=0.7)} &\multicolumn{3}{c|}{Car AP BEV (IoU=0.7)} \\
  Methods & Easy & Moderate & Hard & Easy & Moderate & Hard\\
  \hline
  Baseline                    & 22.08 & 15.64 & 13.00 & 28.56 & 21.08 & 17.48 \\
  Dilation Convolution(2,2,2) & 20.32 & 14.07 & 12.06 & 27.23 & 19.70 & 16.72 \\
  Dilation Convolution(3,3,3) & 17.93 & 12.93 & 10.71 & 24.93 & 17.93 & 15.68 \\
  DCNv2\cite{dcnv2}(n=1)                  & 22.13 & 16.20 & 13.44 & 29.19 & 21.58 & 18.57 \\
  DCNv2(n=2)                  & 20.21 & 15.52 & 13.23 & 29.03 & 21.78 & 18.93 \\
  DCNv2(n=3)                  & 20.21 & 14.58 & 12.24 & 28.12 & 19.79 & 17.11 \\
  RFB\cite{rfb}                         & 21.32 & 15.62 & 12.94 & 28.53 & 21.26 & 18.35 \\
  ASPP\cite{deeplabv2}                        & 22.44 & 16.96 & 14.23 & 29.69 & 22.20 & 19.03 \\
  \textbf{PAC(Ours)}          & 22.71 & 15.73 & 13.05 & 30.55 & 21.85 & 18.56 \\
  \textbf{PAC Module(Ours)}   & \textbf{23.53} & \textbf{17.23} & \textbf{14.33} & \textbf{30.57} & \textbf{22.44} & \textbf{19.07} \\
  \hline
\end{tabular}
\end{center}
\end{table*}


As shown in our experimental results in Table \ref{tab:compare_with_modules}, our proposed PAC module outperforms all other methods and achieves an improvement of +1.59\% AP compare to baseline in the 3D metric with moderate difficulty. In contrast, dilation convolution showed no improvement compared to the baseline. Deformable convolution had a slight improvement with a single layer, but its performance dropped after using more than one DCN layer. We suspect this is because DCN introduces too many parameters to train, making it easier to overfit the training data, especially since KITTI's training set is small. RFB showed roughly the same performance as the baseline, while ASPP showed a fair improvement, especially with hard-difficulty objects. This indicates that far or truncated objects require long-range information to predict their depth accurately.

\subsection{Qualitative Result}
\begin{figure*}[htbp!]
  \centering
  \includegraphics[width=14cm]{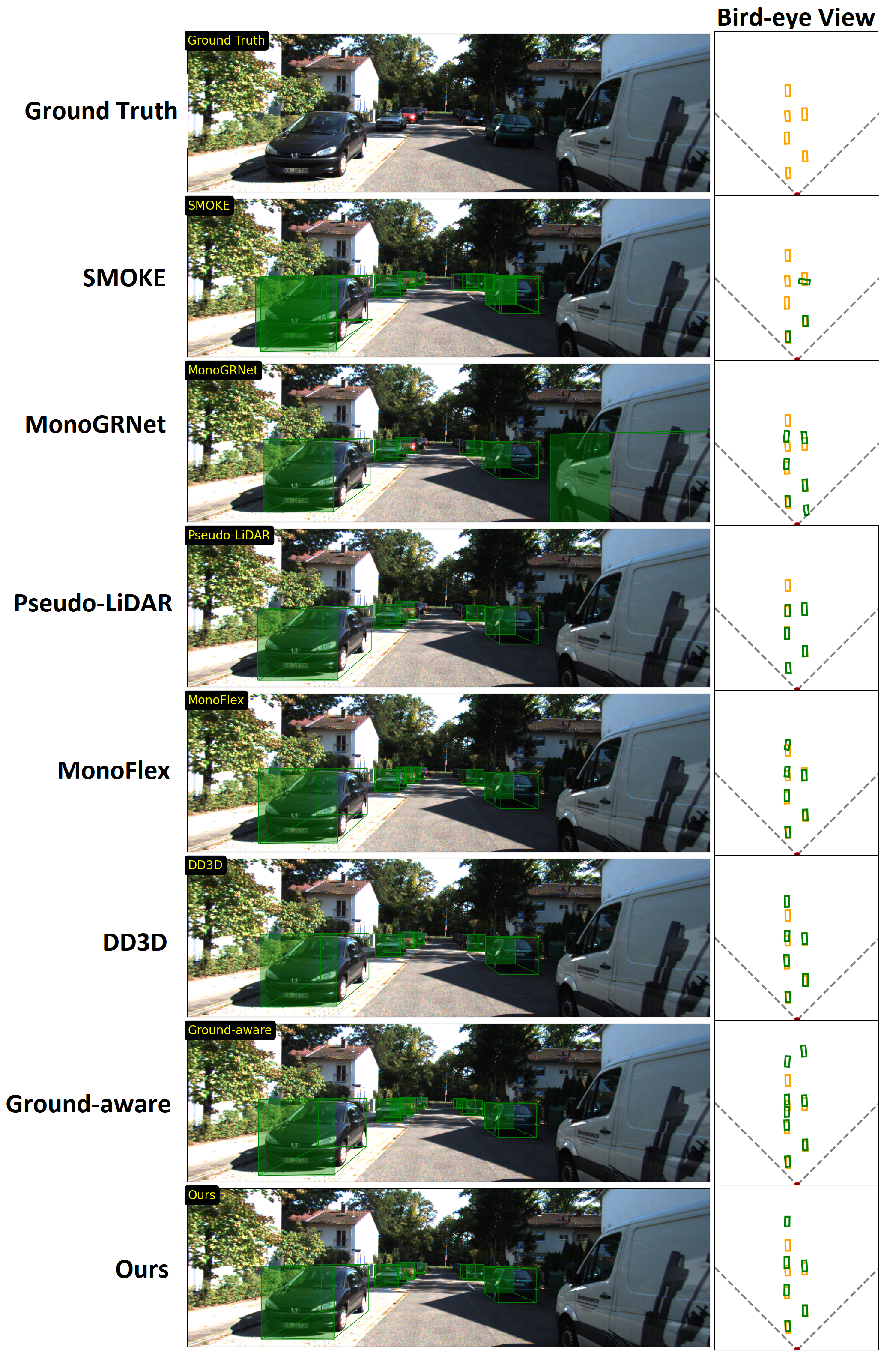}
  \caption{Inference example of 3D object detectors. The left column shows the predicted 3D bounding boxes in the image plane, while the right column displays the bounding boxes projected onto the bird's-eye-view (BEV) plane. The yellow boxes on the BEV represent the ground truth, while the green boxes indicate the predictions. In this experiment, we observe that most methods exhibit inaccuracies in predicting object depth, whereas our proposed method demonstrates more accurate depth estimation.}
  \label{fig:000053}
\end{figure*}

To facilitate a fair comparison of different 3D object detectors, we present some inference outcome example in the KITTI3D validation set, as shown in Figures \ref{fig:000053}. In the figure, the left column shows the predicted 3D bounding box projected onto the image plane. However, to accurately evaluate the 3D box location, we recommend that readers use the bird's-eye-view (BEV) provided in the right column. In the BEV figures, the yellow box represents the ground truth and the green box represents the predicted result.

Based on these results, we can observe some interesting traits of each detector. Firstly, despite the impressive accuracy of Pseudo-LiDAR in 3D box estimation, it can sometimes incorrectly predict the object orientation in pretty obvious cases. This is because Pseudo-LiDAR converts the image to a point cloud, sacrificing some advantages that are only available when the data is in image form. As for MonoFlex and Ground-aware, they perform roughly the same in this experiment, showing keypoint-based and anchor-based method both has potential in 3D object detection. DD3D, on the other hand, tends to generate too many false positives, although their confidence scores are low. This also highlights the advantage of anchor-based methods, where non-maximum suppression is applied to avoid similar issues.

\section{Conclusions}
In this paper, we introduced a novel perspective-aware convolution layer to address the limitations of traditional convolutional kernels in capturing long-range dependencies in images. The PAC module enforces the convolutional kernel to extract features along the perspective lines, making it able to extract perspective-aware features. We integrated the PAC module into a 3D object detector and evaluated its performance on the KITTI3D dataset. The experimental results demonstrate that our approach achieved significant improvements, achieving a 23.9\% AP in the easy difficulty of the dataset and surpassing other 3D object detectors. Our findings highlight the importance of modeling scene clues for accurate depth inference in camera images and the benefits of incorporating perspective information into the neuron network. We believe that our proposed methods have the potential to application in autonomous driving, to enhance 3D object detection accuracy and driving safety.

\bibliographystyle{IEEEbib}
\bibliography{reference}

\end{document}